\documentclass[10pt, a4paper]{article}

\usepackage[final]{lrec2026} 

\usepackage{longtable}
\usepackage{graphicx}
\usepackage{adjustbox}
\usepackage{amssymb}
\usepackage{amsmath}
\usepackage{amsfonts}
\usepackage{subfigure}
\usepackage{subcaption}
\usepackage{array,booktabs,ragged2e}
\usepackage{multirow}
\usepackage{multicol}
\usepackage{latexsym}
\usepackage{arydshln}
\usepackage{pifont}
\usepackage{float}
\usepackage{bm}
\usepackage{xspace}
\usepackage{placeins}
\usepackage{afterpage}

\title{PAREDA: A Multi-Accent Speech Dataset of Natural Language Processing Research Discussions}

\name{Sicheng Jin, Dipankar Srirag, Aditya Joshi} 

\address{University of New South Wales \\
         \{stefan\_zalkoszin.jin, d.srirag, aditya.joshi\}@unsw.edu.au\\}

\abstract{
While modern Automatic Speech Recognition (ASR) systems achieve high accuracy on benchmark corpora, their performance often degrades when there is real-world variability. This work focuses on variability arising due to accented, spontaneous, and domain-specific speech. In particular, we introduce PAper REading DAtaset (PAREDA), a first-of-its-kind multi-accent speech dataset consisting of discussions on academic Natural Language Processing (NLP) papers between speakers with Australian, Indian-English, and Chinese English accents. Each session elicits a spontaneous monologue (a summary of a paper's abstract) and a non-monologue (a question-and-answer session between participants), resulting in a corpus rich with technical jargon and conversational phenomena. We evaluate the performance of SOTA ASR models on PAREDA, analysing the impact of accent mixing and increased speech rate. Our results show that, in the zero-shot setting, models perform worse, confirming the dataset's challenging nature. However, fine-tuning on PAREDA significantly reduces the Word Error Rate (WER), demonstrating that our dataset captures linguistic characteristics often missing from existing corpora. PAREDA serves as a valuable new resource for building and evaluating more robust and inclusive ASR systems for specialised, real-world applications.
 \\ \newline  }

\begin{document}

\maketitleabstract

\section{Introduction}\label{sec:intro}
Automatic speech recognition (ASR) models are increasingly deployed in academic settings such as lecture transcription, workplace meetings, and conference presentations. As these settings become more diverse, ASR systems are required to handle a wide range of linguistic phenomena, including accented speech, dialectal variation, and domain-specific terminology~\cite{mehrish2023review}. While recent ASR models achieve strong performance on general-domain monologue speech for mainstream accents, studies have shown that state-of-the-art ASR models such as Whisper~\cite{pmlr-v202-radford23a} exhibit degraded transcription performance for non-mainstream English accents, including Nigerian English (en-NG) and Indian English (en-IN), when compared to mainstream American English (en-US)~\citeplanguageresource{eisenstein2023md3}. This evaluation is for general-domain dialogue.

Academic conversations are inherently domain-specific and frequently involve speakers with diverse linguistic backgrounds, as in the seminal AMI meeting corpus~\citeplanguageresource{kraaij2005ami}. To our knowledge, there is currently no publicly available resource designed to evaluate ASR performance on domain-specific academic speech involving non-mainstream English accents. In this work, we describe the creation of PAREDA (which stands for PAper REading DAtaset), a small-scale multi-accent speech dataset comprising 3.9 hours of recorded audio. The dataset consists of conversations between pairs of speakers with non-mainstream English accents, specifically Australian, Chinese, and Indian English. Speakers exhibit varying levels of expertise in natural language processing, ranging from students to researchers. The conversations center on discussions of NLP research papers sourced from the ACL Anthology~\footnote{\href{https://www.aclanthology.org/}{https://www.aclanthology.org/}}. This design reflects realistic academic interactions, where the use of technical terminology varies according to a speaker’s familiarity with the subject matter, resulting in differing densities of domain-specific vocabulary across utterances. Our evaluation on PAREDA highlights that ASR models like Whisper~\cite{pmlr-v202-radford23a}, Phi-4~\cite{abouelenin2025phi}, and CrisperWhisper~\cite{wagner2024crisperwhisper} are able to achieve SOTA accuracy, but certain conditions of the input audio have to be met. The domain choice (academic NLP discussions) and its intersection with accent variability make PAREDA a useful dataset for future research.

\begin{figure*}
    \centering
    \includegraphics[width=1\linewidth]{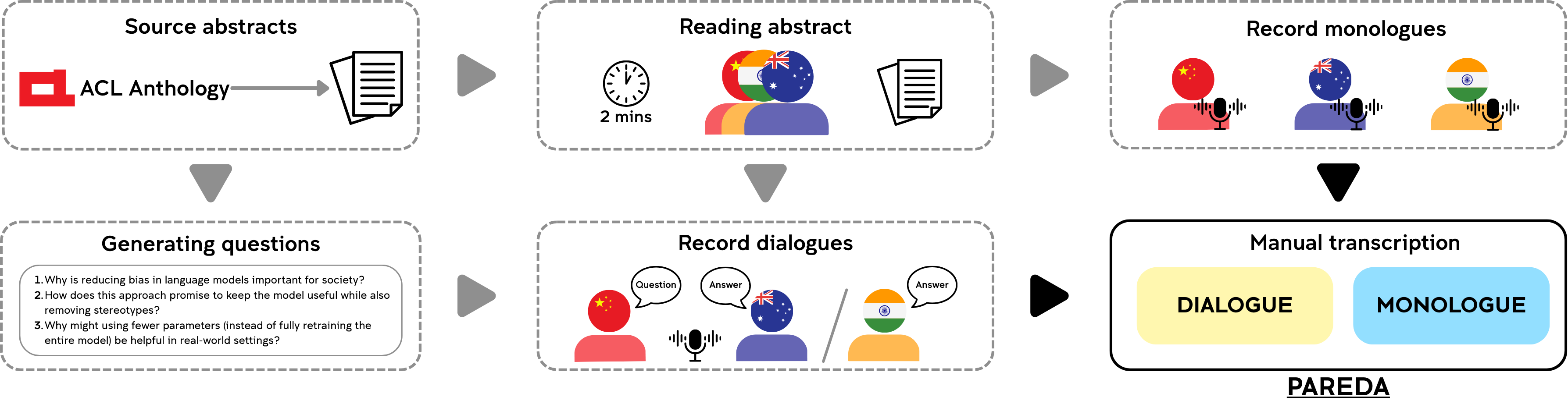}
    \caption{Methodology for dataset collection}
    \label{fig:intro-collect}
\end{figure*}

\section{Data Collection}\label{sec:data-collect}

We collect data of speech recordings, using the methodology illustrated in Figure~\ref{fig:intro-collect}. We use NLP research papers collected from ACL Anthology to create our dataset. We cover \textit{three} locales, with \textit{one} participant from each locale. Due to the nature of this study, we limit our participant group to \textit{one} speaker per accent. Each speaker is given 21 papers for the elicitation exercise.

\subsection{Speakers and Prompts}\label{sec:elicit}

We conduct elicitation with three participants, one for each locale. The three locales covered in this dataset are: Australian (en-AU), Indian (en-IN), Northern Chinese (en-ZH). We did not collect American (en-US) samples as there is already an excessive amount of en-US speech samples available in other datasets, and the models we use have already exposed to such audio extensively during training. All speakers have experience in NLP research and were willing to engage in recording for this research. The Indian speaker is an NLP lecturer with extensive research expertise in the academic field, and the other two speakers are NLP research students. All participants are also above 18 years of age, with native or superior proficiency in English\footnote{We determine the proficiency of non-native speakers using English language tests like IELTS.}. We prepared a corpus of 21 NLP research papers on the ACL Anthology website which, according to their focus, can be broadly categorised as: (A) NLP in applied linguistics, (B) NLP in linguistic research, (C) Mitigating NLP bias, (D) NLP in historical and cultural linguistics.

\subsection{Recording}

The speech samples are either collected online or in person in an indoor environment in a closed meeting room. During each recording session, the participant is given two minutes to read the abstract. If the abstract was found to be non-informative, the participant is allowed to read the full body and summarise the paper. Following this, we also record discussion sessions involving a host asking questions. These questions are specific to each paper, and were prepared to elicit utterances and guide conversation between the host and the participant. This discussion session lasts for upto five minutes. Figure~\ref{fig:intro-collect} demonstrates an example of this procedure. We manually segment the audio files to less than 30 seconds to ensure the audio is not split abruptly mid-sentence. This is done as some models experimented in this paper only support audio samples of less than 30 seconds. The final audio files are saved in the .wav format. We denote the summarisation session of the speech sample as \textit{monologue}, and the discussion sessions as the \textit{dialogue}. The \textit{dialogue} subset does not involve full-duplex speech; rather, it consists of structured question–answer exchanges.

\subsection{Transcription}

We first generate raw transcriptions using the ASR model CrisperWhisper \cite{wagner2024crisperwhisper}, which are then manually post-edited. As the models evaluated in this study default to American English spelling conventions, the transcriptions are further standardised to follow the same convention. To assess inter-annotator reliability, an independent annotator post-edited 10 randomly selected transcriptions from the \textit{dialogue} subset. Agreement between annotators was quantified using WER, computed after standard text normalisation. The resulting mean symmetric WER was 2.77\%, indicating high agreement between annotators.

\subsection{Dataset Statistics}\label{sec:stats}
\begin{table}[t] 
    \centering
    \resizebox{\columnwidth}{!}{%
    \begin{tabular}{lccc}
    \toprule
     & en-AU & en-IN & en-ZH \\
    \midrule
    Monologue & 16:19 & 36:22 & 64:24 \\
    Non-Monologue  & 75:21 & 41:58 & -- \\
    \hdashline
    Total     & 91:40 & 78:20 & 64:24 \\
    \bottomrule
    \end{tabular}}
        \caption{Speech duration (minutes:seconds) by accent and interaction type. Non-monologue duration reflects respondent speech only; the en-ZH speaker serves as the host during dialogues and is therefore not included as a dialogue respondent.}
    \label{tab:dataset-setup}
\end{table}

Table~\ref{tab:dataset-setup} summarises the distribution of speech duration across accents and interaction types. Monologue recordings were collected from all three accents. Non-Monologue recordings consist of structured question–answer exchanges in which speakers with en-AU and en-IN accents act as respondents, while the en-ZH speaker serves exclusively as the host and questioner. Consequently, Non-Monologue duration is reported only for en-AU and en-IN accents. The dataset contains 20 en-AU, 23 en-IN, and 39 en-ZH monologue samples, as well as 50 en-AU and 28 en-IN dialogue samples. We split the dataset into training, validation and test with 80:10:10 ratio.



\section{Experiment Setup}\label{sec:setup}


We fine-tune four model sizes of Whisper in a two-stage training procedure. In the first stage, the models are fine-tuned on GLOBE, a large-scale multi-accent English speech corpus, to improve robustness to accented speech. All GLOBE fine-tuned models are trained for 5,000 steps with an initial learning rate of 1 e-5 and a warmup ratio of 10\%. Model checkpoints are evaluated every 500 steps, and the resulting models are subsequently evaluated on our dataset to establish baseline performance prior to in-domain adaptation. In the second stage, the GLOBE-tuned models are further fine-tuned on our collected dataset to adapt the models to domain-specific academic speech. As the in-domain dataset is substantially smaller than GLOBE, we increase the evaluation frequency and employ early stopping to mitigate overfitting. OpenAI's Whisper API, Microsoft's Phi-4 and CrisperWhisper are also used for evaluation, as these are leading models on Hugging Face's Open ASR Leaderboard at the time of experiment. CrisperWhisper was selected as it is specifically tailored for producing verbatim transcripts for casual speech. Whisper API used Whisper Large as the model, while Phi-4 and CrisperWhisper were obtained from HuggingFace. Experiments requiring no fine-tuning were carried out using one Nvidia A100 GPU and all fine-tuning experiments were performed on one Nvidia Tesla Volta GPUs. The total training time for both stages was approximately 70 hours.


\section{Results}\label{sec:results}

\begin{table*}
    \centering
    \begin{tabular}{llcccccc}
    \toprule
    {Condition} & {Model} & {en-AU} & {en-AU/ZH} & {en-IN} & {en-IN/ZH} & {en-ZH} & {en-US}\\
    \midrule
    Normal          & Whisper API & 18.21 & 15.04 & 9.56  & 10.62 & 15.04 & 3.91 \\
                    & Phi4        & 8.62  & 8.69  & 8.96  & 9.15  & 8.61  & 3.82 \\
                    & CrisperWhisper& 5.10  & 4.29  & 4.08  & 4.66  & 4.38  & 3.97\\
    \midrule
    1.5x Speed          & Whisper API & 25.98 & 23.56 & 14.76 & 16.49 & 20.76 & -\\
                        & Phi4        & 20.77 & 22.37 & 16.16 & 21.23 & 22.98 & -\\
                        & CrisperWhisper& 25.57 & 25.74 & 17.05 & 19.05 & 22.24 & -\\
    \midrule
    -10dB Noise & Whisper API & 22.51 & 19.11 & 14.65 & 15.40 & 21.10 & -\\
                         & Phi4        & 14.12 & 12.80 & 10.95 & 13.18 & 14.98 & -\\
                         & CrisperWhisper& 10.87 & 12.41 & 9.51  & 17.94 & 27.67 & -\\
    \bottomrule
    \end{tabular}
    \caption{WER (\%) Benchmark Across ASR Architectures Under Varied Linguistic and Environmental Conditions. Note that the last column, en-US, is taken from the \href{https://huggingface.co/spaces/hf-audio/open_asr_leaderboard}{Hugging Face Open ASR Leaderboard} on the Librispeech-other dataset.}
    \label{tab:combined_wer}
\end{table*}

We address the following questions in our evaluation using PAREDA and other datasets: (a) Would mixed accents worsen the performance of ASR models? (b) Would speech speed influence the performance of ASR models? (c) Can models fine-tuned on non-casual speech perform equally well on casual speech? Table~\ref{tab:combined_wer} details the performance of these three models under different speech accent and scenario combinations using word error rate (WER) as the metric. For our dataset, there are two major accents: Australian and Indian, designated by en-AU and en-IN; while en-ZH represents Northern Chinese, an accent added into some speech samples of the two major accents. Since en-ZH is used as a condition accent to help assessing other accents, no separate en-ZH samples are evaluated.

\subsection{Mixed Accents}\label{sec:non-typical}


Table~\ref{tab:combined_wer} shows that, model-wise, CrisperWhisper achieves the highest WER across all speech accents and speech types, with the overall lowest WER being 4.08\% on Indian speech samples, while Phi-4 performs slightly lower. Whisper API exhibits the lowest scores, producing the highest WER of 18.21\% on Australian samples. However, there is no clear evidence of how introducing en-ZH influences WER. We observe that for Phi-4, the WER for both accents are higher when adding en-ZH: Australian WER increases from 8.62\% to 8.69\% and Indian WER increases 8.96\% to 9.15\%; however, for the other two models, Australian WER drops considerably with en-ZH. The en-ZH monologue samples score relatively similar to Australian ones, with the whisper WER being slightly lower. Indian samples typically result in lower WER, although this trend does not apply to all models, as observed in the 8.96\% result from Phi-4, which is higher than both Australian samples 8.62\% and 8.69\%. For the other two models, it is clear that Indian monologue samples performs better than Australian monologue ones, while Australian non-monologue samples might show lower WER than the Indian counterparts, as in Whisper API results, or higher WER, as in CrisperWhisper results. Similarly, adding another accent such as en-ZH does not result in a noticeable shift in the WER. Whether the mixed speech would score a higher or lower WER relates to the model used. In our example, the Indian non-monologue samples had a higher WER than the monologue ones, however this pattern reverses of Australian samples. What did not alter is the overall pattern of WER of the models, where CrisperWhisper performs significantly better than the other two.

\subsection{Noise Robustness}\label{sec:speed}
We see the opportunity to evaluate for robustness to noise by synthetically adding variations. We consider two additional configurations: increasing the speed of the audio and adding background noise. When we accelerate the speech to 1.5 times faster or add a -10dB background white noise, recognition accuracy drops sharply for every model, as shown in Table~\ref{tab:combined_wer}. For accelerated tests, Whisper API’s error rate rises from about 18\% to 26\% on the Australian samples and from 10\% to 15\% on the Indian samples; Phi-4 shows a similar pattern, jumping to roughly 21\% and 16\% respectively. CrisperWhisper experienced the worst degradation where its Australian error climbs significantly to 26\%, and its Indian error almost quadruples to 17\%. For noise tests, the error rate did not increase to the extent of the accelerated samples, however, still visibly worse than the original normal test results. Whisper's WER rises to 22.5\% for Australian samples and from 10\% to 14.6\% on Indian samples. The other models follow the same pattern. In general, faster delivery or excessive background noise hinders the advantages of the strongest model and narrows the gap between systems. This suggests that none of the three architectures has learned a truly tempo-invariant acoustic representation, and that further speed-augmentation or specialised front-end processing may be required to restore their baseline performance. Similar results are observed for the introduction of background noise. For all three models, combining fast speech with accent mixing still produces error rates in the low-to-mid 20\% range, roughly double the figures observed when only the extra speaker is present. Despite the overall degradation, we can observe in Figure~\ref{tab:combined_wer} that Whisper API remains the least robust, CrisperWhisper the most accurate, and Phi-4 sits in between. The fact that accent diversity has so little impact once speed increases implies that speaking-rate variation is a dominant constraint. Future work should therefore prioritise systematic tempo-augmentation and dynamic time-warping techniques, as they appear more promising than simply expanding accent coverage for enhancing real-world resilience.



\begin{table*}[t!]
    \centering
    \begin{tabular}{lcccc}
    \toprule
    & \multicolumn{4}{c}{{Whisper Model Size}} \\
    \cmidrule(l){2-5}
    {Fine-Tuning Stage} & {Tiny} & {Small} & {Medium} & {Large} \\
    \midrule
    Baseline (Not Fine-tuned) & 22.20& 15.03& 13.46& 15.39\\
    Stage 1 (GLOBE-tuned)     & 23.95& 18.01& 15.84& 16.41\\
    Stage 2 (PAREDA-tuned)    & 12.85& 6.68& 4.53& 4.87\\
    \bottomrule
    \end{tabular}
    \caption{WER Comparison when fine-tuning Whisper with/without PAREDA.}
    \label{tab:q3-variety}
\end{table*}

\subsection{Cross-variety Evaluation}\label{sec:cross}

We also conduct a cross-variety evaluation to examine whether models fine-tuned on non-casual datasets can generalize to PAREDA’s casual, technical speech, and whether domain-specific fine-tuning provides significant gains. Table~\ref{tab:q3-variety} compares performance across three stages: zero-shot baseline, GLOBE-tuned, and PAREDA-tuned. In the baseline stage, Whisper Medium achieved the lowest WER (13.46\%), notably outperforming the Large model, while the Tiny version performed worst at 22.20\%. Surprisingly, Stage 1 fine-tuning with GLOBE resulted in "peculiar" results, as WER actually increased across all model sizes—ranging from a 1.02\% rise for Large to 2.98\% for Small. Following this, we further fine-tuned the models using PAREDA until WER and validation loss stagnated, as detailed in Section~\ref{sec:setup}. Evaluating these Stage 2 models against the PAREDA test set revealed substantial improvements, with WER for every version dropping by at least 10\% compared to previous stages. These findings demonstrate that while broad multi-accent datasets like GLOBE provide breadth, they may not contribute to improved recognition in specialized professional fields. This highlights the necessity of incorporating casual, domain-specific speech during training to help ASR models comprehend the distinct phonological nuances found in expert-level discourse.

\subsection{Per-Accent Evaluation}\label{sec:peraccent}

\begin{figure}
    \centering
    \includegraphics[width=1\linewidth]{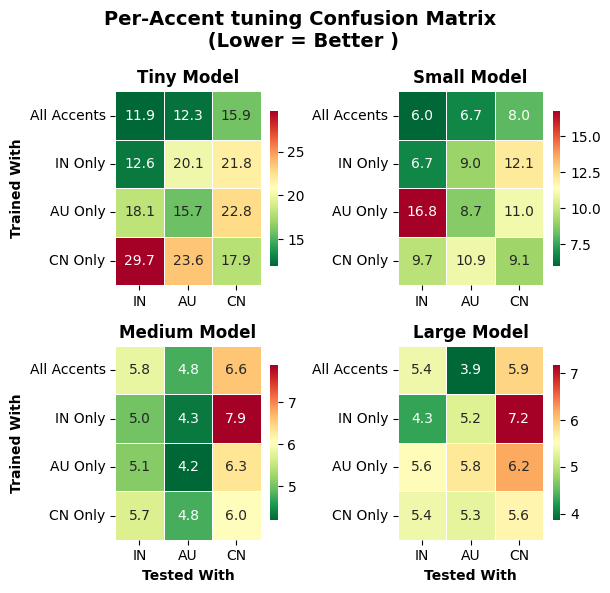}
    \caption{Per-Accent Tuning Results}
    \label{fig:peraccent-1}
\end{figure}

\begin{figure}
    \centering
    \includegraphics[width=1\linewidth]{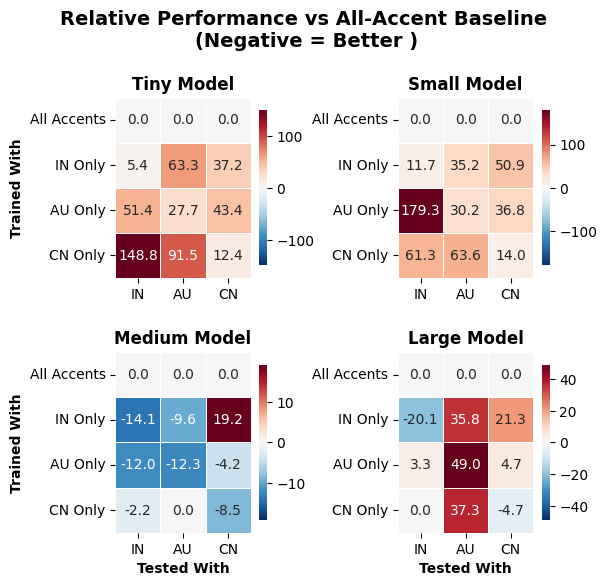}
    \caption{Per-Accent Relative Results}
    \label{fig:peraccent-2}
\end{figure}

We then perform a per-accent evaluation, fine-tuning GLOBE-tuned models on individual PAREDA accent subsets to observe if targeted tuning reduces the Word Error Rate (WER) for specific accents. We maintained the Stage 1 configuration, using an early stopping patience of 3 and a 1e-5 learning rate. Evaluation steps were set to 250 for Tiny/Small models and 50 for Medium/Large versions. Figures \ref{fig:peraccent-1} and \ref{fig:peraccent-2} present the raw WER and relative performance against the all-accent baseline, with rows representing the tuning accent and columns representing the test accent.

For Tiny and Small models, per-accent tuning generally worsened performance compared to the all-accent baseline. For instance, the Indian-tuned Tiny model's WER rose from 11.9 to 12.6 on Indian samples, while cross-accent testing (e.g., tuning on Indian/Australian and testing on Chinese) resulted in significantly higher errors. Conversely, per-accent tuning benefited larger models, particularly Medium, where WER decreased across all accents. The Large model showed a contradictory pattern for Australian samples, where tuning on its own accent caused the highest WER increase, though other accents saw modest improvements. These results suggest that while phonology is a factor, model complexity and contextual speech characteristics—such as technical vocabulary and filler words—are equally decisive in ASR performance.

\subsection{Qualitative Analysis}

Beyond aggregate metrics, we performed a linguistic error analysis on the Whisper-tiny model's PAREDA test performance. Using NLTK and WordNet, we categorised errors at the word level to understand how specific accent conditions influence ASR failure. In order to conduct the analysis, we utilised a sequence alignment approach using Python's \verb|difflib.SequenceMatcher|\footnote{\url{https://docs.python.org/3/library/difflib.html}} to identify three types of errors: (1) {substitutions} (one-to-one word replacements), (2) {deletions} (words present in reference but missing in prediction), and (3) {insertions} (words present in prediction but absent in reference). To avoid double-counting, substitutions were identified first through sequence alignment, and only words not involved in substitutions were considered for deletion/insertion analysis. 

Each error word was automatically categorised using NLTK's part-of-speech tagging, stopword identification, and WordNet's semantic taxonomy. This approach eliminated manual categorisation bias and provided consistent, reproducible linguistic classifications across all 488 unique error instances identified in the 77-sentence test set. The analysis revealed 84 unique substitutions, 196 deletions, and 208 insertions, totalling 734 errors and 488 distinct occurrences.

The top three semantic classes remain the same, but their relative weight has shifted: (A) \textbf{Function words}, which mostly serve grammatical purposes such as \textit{`a', `do', `can'} now account for 241 tokens (29.3\%), still the single largest source of error. Deletions of articles (\textit{a}, \textit{the}) and prepositions dominate, confirming the model’s difficulty with reduced, unstressed items; (B) \textbf{Short filler tokens} (\textit{uh}, \textit{um}, single letters) rose sharply to 131 errors (16\%), with a striking 11.9 errors per unique filler, by far the densest error class; \textbf{Communication-related terms}, including many NLP words such as \textit{`prompting', `slang', `dialogue'} contribute 54 errors (6.6\%). We present additional examples fo this class in Table\ref{tab:transcript_samples}. Smaller but noteworthy movements are observed in the following cases. \textbf{Unclassified/OOV items} climbed to 49 tokens, many are the result of hallucinated insertions rather than deletions, including non-existent words such as \textit{`afimouth'} and \textit{`dissist`,} hinting at domain‑dependent lexical over‑generation, and that accented pronunciations still confuse the ASR. Similarly, \textbf{morphological‑technical terms} remain rare: only one token (``tokenisation'') was mis‑recognised, which was consistent with the previous finding that complex morphology is handled reasonably well. Function‑word errors break down into deletions 40\%, insertions 44\%, substitutions 16\% (percentages of the 241 tokens). The prevalence of function‑to‑function substitutions including \textit{a$\rightarrow$the} and \textit{will$\rightarrow$for} signals acoustic confusion among weak syllables, though their absolute counts have increased with the larger error pool.

\begin{table*}[h!]
    \centering
    \renewcommand{\arraystretch}{1.5} 
    
    \begin{tabularx}{\linewidth}{X X@{}} 
        \toprule
        \textbf{Reference} & \textbf{Prediction} \\ 
        \midrule
        
        \dots just a parser may not be sufficient and there will be other tools such as stemmers \dots 
                           & \dots just a \textbf{person} may not be sufficient and there will be other tools such as \textbf{standard} \dots \\
        
        \addlinespace[0.5em]
         \dots it is a low resource language \dots 
         & \dots it is a \textbf{only source} language \dots \\
        
        \midrule
        
        \dots how many n grams they get right \dots 
          & \dots how many \textbf{enzymes} they get \textbf{ripe} \dots \\
        
        \bottomrule
    \end{tabularx}
    \caption{Examples of ASR transcription errors. Mismatches between the Reference (Ground Truth) and Model Prediction are highlighted.}
    \label{tab:transcript_samples}
\end{table*}



\begin{table*}[t]
    \centering
    \small
    \begin{tabular}{lcccccr}
        \toprule
        Category &
        \shortstack{Total\\Words} &
        \shortstack{Words in\\Test Set} &
        \shortstack{Total\\Appearances} &
        \shortstack{Total\\Errors} &
        \shortstack{Words w/\\Errors} &
        \shortstack{Avg. Error\\Rate}\\
        \midrule
        Content (NLP) & 54 & 41 & 117 & 35 & 26 & 45.80\% \\
        Grammatical   & 46 & 45 & 966 & 62 & 25 & 7.57\% \\
        \bottomrule
    \end{tabular}
    \caption{Linguistic Error Analysis: Technical vs. Functional Vocabulary}
    \label{tab:452-compare-table}
\end{table*}

Our targeted word‑list comparison covers 54 NLP terms and 46 functional words, as seen in Table \ref{tab:452-compare-table}. Some words are not observed in the test set, and those are not counted towards the results. These include: (A) \textbf{Domain words}: 45.8\% average error rate (35 errors in 117 appearances); (B) \textbf{Grammatical words}: 7.6\% (62 errors in 966 appearances). The ratio {6$\times$} confirms that technical vocabulary is still markedly harder for the model to capture.

These patterns reveal two primary pathways for ASR failure in multi-accent, technical discourse. First, a ``prosodic deletion'' pathway affects function words and fillers, where accent-conditioned vowel reduction leads to the omission of unstressed items like articles and prepositions. Second, an ``acoustic mis-modeling'' pathway targets low-frequency specialist vocabulary. For the 54 NLP terms analysed, the average error rate was 45.8\% which is six times higher than the 7.6\% rate for grammatical words. This {6$\times$} disparity suggests that uncommon phoneme sequences combined with accent shifts overwhelm the recognizer, regardless of its underlying lexical knowledge. Addressing these challenges requires a dual-track approach: prosody-aware modeling to capture variant pronunciations of function words and lexical specialisation to boost the recognition of domain-specific terminology.

\section{Related Work}\label{sec:relwork}

SOTA ASR systems often achieve high accuracy on standard benchmarks but exhibit significant performance degradation across diverse speaker groups, a phenomenon known as ASR bias~\cite{mehrish2023review, feng2024towards}. Variations in pronunciation, accent, and intonation remain persistent challenges, often rooted in the lack of diversity within training corpora~\cite{basak2023challenges, feng2024towards}. Recent efforts to address this include the GLOBE corpus, which provides read speech across 164 accents, though it lacks spontaneous conversational nuances~\citeplanguageresource{wang2024globe}. Conversely, the Multi-Dialect Dataset of Dialogues (MD3) offers task-oriented conversational data but focuses on general-domain tasks like guessing games~\citeplanguageresource{eisenstein2023md3}. Datasets of meeting corpora either rely on general topics~\citeplanguageresource{mccowan2005ami} or use TTS to generate audio~\citeplanguageresource{lee2023dailytalk}. PAREDA addresses a critical gap these corpora overlook: the intersection of dialectal variation and domain-specific, technical vocabulary in spontaneous speech. Unlike read-speech datasets or general-knowledge dialogues, PAREDA targets the challenge of accented speech and technical jargon, which is known to cause robustness issues even in SOTA models like Whisper~\cite{jain2024exploring, radford2023robust}.

\section{Conclusion}\label{sec:concl}

We introduced PAper REading DAtaset (PAREDA), a novel multi-accent speech dataset comprising spontaneous monologues and non-monologues from Australian, Indian, and Chinese English speakers discussing technical NLP research papers. The design of PAREDA elicits natural, casual speech laden with domain-specific jargon, creating a challenging and realistic testbed for modern ASR systems. Our experiments demonstrate that a SOTA model, Whisper, yields unsatisfactory Word Error Rates (WER) on the PAREDA corpus in a zero-shot setting, confirming that dialectal and domain-specific speech remains a significant hurdle~\cite{mehrish2023review}. The analysis revealed particular difficulties with domain-specific vocabulary, accent mixing, and variations in speaking rate. However, we also showed that finetuning the Whisper model, even with the limited data in PAREDA, leads to considerable performance improvements. This finding aligns with research in other specialised, low-resource domains, such as child speech recognition, where finetuning is a highly effective strategy~\cite{jain2024exploring}. Future work should focus on expanding the PAREDA dataset to include more speakers and a wider variety of global Englishes. PAREDA can be used to investigate the propagation of ASR errors into downstream NLP tasks and serve as a crucial benchmark for developing novel bias mitigation techniques aimed at creating more inclusive speech recognition technologies~\cite{feng2024towards}.

\section{Bibliographical References}\label{sec:reference}

\bibliographystyle{lrec2026-natbib}
\bibliography{refer}

\section{Language Resource References}
\label{sec:langref}
\bibliographystylelanguageresource{lrec2026-natbib}
\bibliographylanguageresource{languageresource}

\end{document}